\documentclass[runningheads]{llncs}
\usepackage{graphicx}
\usepackage{hyperref}
\usepackage{times}
\usepackage{xcolor}
\usepackage{amsmath,amssymb,bbm}
\usepackage{enumitem}
\usepackage{booktabs}  
\newcommand{\R}{\mathbb R}

\newcommand{\E}{\mathbb E}
\newcommand{\Pbb}{\mathbb P}
\newcommand{\one}{\mathbbm{1}}

\newcommand{\Fcal}{\mathcal{F}}

\newcommand{\Lcal}{\mathcal{L}}

\newcommand{\blind}{1}

\begin{document}
\title{Convergence of a L2 regularized Policy Gradient Algorithm for the Multi Armed Bandit}
\titlerunning{Regularized policy gradient for MAB}
%

\if1\blind
{
\author{
\c{S}tefana-Lucia Ani\c{t}a\inst{1}\orcidID{0000-0003-3369-9551} \and
Gabriel Turinici\inst{2}\orcidID{0000-0003-2713-006X} 
}%
\authorrunning{S. Anita and G. Turinici}
%
\institute{``Octav Mayer'' Institute of Mathematics of the Romanian Academy\\ Bd. Carol I 8, Ia\c{s}i 700505, Romania \\
		\email{stefi\_anita@yahoo.com}\\
\and
CEREMADE, \\  Universit\'e Paris Dauphine - PSL, CNRS, Paris, France \\
		\email{gabriel.turinici@dauphine.fr}\\
		\url{https://turinici.com} }
}\fi
\if0\blind{
\author{anonymized version\institute{anonymous institute}}
}\fi

\maketitle              
\begin{abstract}
Although Multi Armed Bandit (MAB) on one hand and the policy gradient approach on the other hand are among the most used frameworks of Reinforcement Learning, the theoretical properties of the policy gradient algorithm used for MAB have not been given enough attention. We investigate in this work the convergence of such a procedure for the situation when a $L2$ regularization term is present jointly with the 'softmax' parametrization. We prove convergence under appropriate technical hypotheses and test numerically the procedure including situations beyond the theoretical setting. The tests show that a time dependent regularized procedure can improve over the canonical approach especially when the initial guess is far from the solution.
\end{abstract}
\section{Introduction}

Supported by impressive practical applications including 
game play (e.g., Go \cite{silver_rl_mastering_go_2016}, computer games \cite{mnih_human_level_rl_2015}), autonomous car driving \cite{bojarski2016end_rl_cars},  ChatGPT \cite{openai2024gpt4}, healthcare \cite{RL_medecine1,RL_medecine2}, recommender systems \cite{rl_recommender22} etc.,  the Reinforcement Learning is a promising area of active research today. Standing out among Reinforcement Learning frameworks, the Multi Armed Bandit (MAB in the sequel) \cite{contextual_mab,slivkins_introduction_mab_2019} has been extensively used both for theoretical investigations and for applications.
We will focus here on a specific procedure, the softmax parameterized policy gradient as in \cite[section 2.8 and chap. 13]{sutton_reinforcement_2018}. We investigate its convergence in presence of $L2$ regularization\footnote{In the
machine learning literature the regularization considered here is denoted $L2$  
while in the mathematics literature the $L^2$ notation is more often used; we use  $L2$ throughout the text but both mean the same thing.} and numerically explore the performance of this regularized framework.

The plan of the paper is as follows: in the rest of this section we briefly review the literature while in section \ref{sec:notations} we give the first notations and definitions. Then in section \ref{sec:cv_proof} we prove the convergence under some technical hypotheses, followed in section \ref{sec:numerical} by some numerical tests that confirm the theoretical results and also go beyond it to regimes not covered by the theory. We close with a discussion in section~\ref{sec:conclusion}.

\subsection{Brief literature review}

The policy gradient algorithms have shown impressive results  for applications in reinforcement learning but it has been long recognized that some corrections are necessary to improve convergence; several well known procedures implementing such corrections are the log-barrier 
penalized REINFORCE algorithm \cite{reinforce_cv_proof_Kim_Boyd_2021}, 
trust-region policy optimization TRPO \cite{trpo15} and the proximal policy optimizations (PPO, the OpenAI's default reinforcement learning algorithm); all use a form of regularization, i.e. all seek to limit and control the policy updates by various methods. In this general setting we will focus here on a different type of regularization and will most specifically talk about Multi Armed Bandits.

While the policy gradient algorithms show interesting numerical performance, the theoretical investigations of the convergence for the MAB have only recently witnessed important advances. In \cite{fazel_global_2018} it is proven that stochastic gradients procedures  converge with high probability for the general situation of linear quadratic regulators while Agarwal et al. gave in 
\cite{pmlr_v125_agarwal20a}  theoretical results under the general framework of Markov processes and specifically proved the convergence under different policy parameterizations; on the specific case of softmax parameterization that we analyze here, they examine three algorithms addressing this issue. The initial approach involves straightforward policy gradient descent on the objective without alterations. The second method incorporates entropic regularization to prevent the parameters from growing excessively, thereby ensuring sufficient exploration. Lastly, they investigate the natural policy gradient algorithm and demonstrate a global optimality outcome independent of the distribution mismatch coefficient or dimension-specific factors. Recall that in contrast we study here the softmax parameterization with $L2$ regularization.

In a very recent  paper \cite{bhandari_global_2024} published online just months ago (at the time of writing)
J. Bhandari and D. Russo discuss the softmax parametrization but focus on (we cite) "an idealized policy gradient update with access to exact gradient evaluations".  As a distinction, we will focus here on the non-exact gradient (which is the one usually implemented) but at the price of stronger hypotheses.
Yet in another state-of-the-art research
\cite{mei_global_2020} the authors make three contributions; first they establish that, when employing the true gradient (i.e., without the stochasticity), policy gradient with a softmax parametrization converges at a rate of $O(1/t)$. Then 
they examine entropy-regularized policy gradient and demonstrate its  accelerated  convergence rate. Finally, by integrating the aforementioned outcomes they describe  the mechanism through which entropy regularization enhances policy optimization.

Finally, some other relevant works include \cite{wang2019neural} that study more specifically the situations when deep neural networks are used, while \cite{zhang_global_2020} investigate  the infinite-horizon setting with discounted factors through a new variant that uses a random roll-out horizon for the Monte Carlo estimation.

On a more general theoretical view, as mentioned earlier, our focus is on softmax parameterized policy gradients with $L2$ regularization. 
We will employ arguments similar to that used for the convergence of general stochastic gradient descent (as developed from the initial proposal of Robbins and Monro \cite{robbins_stochastic_1951}). 
A good book on this subject is \cite{chen_stochastic_2002} while 
recent works giving information on the convergence of the SGD for non-convex functions are  \cite{sgd_conv_non_cx20,mertikopoulos_almost_2020}; for short 
self-contained proofs see \cite{gabriel_turinici_convergence_2023,anita2024convergence}.
Note
that classical SGD convergence results as in \cite[Thms 1.2.1 or 1.3.1]{chen_stochastic_2002}
need several hypotheses, for instance the uniqueness of the critical point (here not true), some boundedness conditions (here without any regularization the optimal $H$ will have infinite values), a convenient Lyapunov functional (the obvious one has degenerate directions in this case),
some boundedness for the trajectories~\cite{mertikopoulos_almost_2020} 
and so on. Nevertheless, this will still constitute the basis of our work that puts together estimations and proofs from the literature that were not invoked in this setting before.

\section{The softmax parameterized policy gradient Multi Armed Bandit with $L2$ regularization} \label{sec:notations}

We describe here the softmax parameterized 
 Multi Armed Bandit
policy gradient algorithm to which we add a $L2$ regularization term.
For a description of the original  Multi Armed Bandit (MAB) we refer to \cite{sutton_reinforcement_2018}.
In the classical Multi Armed Bandit problem, we have \(k\) arms indexed by \(a\) where \(a = 1, 2, \ldots, k\).
Each arm \(a\) has an associated reward distribution with mean 
$q_*(a)$. A case often considered is when the reward is  normally distributed with mean 
$q_*(a)$ and variance $\sigma(a)^2=1$ (see later for our hypotheses on $R$ which are more general). 
At each time step \(t\), an agent selects an arm $A_t$ and observes a reward 
$R_t \sim R(A_t)$ sampled from the distribution of the selected arm $A_t$.
The goal is to maximize the cumulative reward over a fixed number of time steps or iterations.

In the policy gradient algorithm with softmax parametrization, the agent maintains a parameterized policy \(\Pi_{H}\), where $H$ is a parameter vector called 'preference vector'.
The preference vector $H$ defines the probability $\Pi _H(A)$ to act on the arm $A$ through the softmax mapping~:
\begin{equation}\Pi _H(A)={\frac{e^{H(A)}}{\sum _{a=1}^ke^{H(a)}}}. 
\label{eq:definition_piha}
\end{equation}

The MAB with regularization is formulated as finding the  optimal preference vector $H\in \mathbb{R}^k$ solution to~:
\begin{equation}
\text{maximize}_{H \in \R^k} \Lcal_\gamma(H),
\label{eq:problem_mab_as_maximization}
\end{equation}
where the functional $\Lcal_\gamma$ is defined as~:
\begin{equation}
\Lcal_\gamma(H) :=\mathbb{E}_{A \sim \Pi_H} \left[ R(A) - \frac{\gamma}{2}\| H\|^2\right].
\label{eq:functional_to_min}
\end{equation}
Here $A \sim \Pi_H$ means that $A$ is sampled from the discrete law $\Pi_H$; $\gamma$ is a positive constant that is seen as a $L2$ regularization coefficient. For convenience, we will sometimes omit the $\gamma$ 
in the notation and  write only
\begin{equation}
\Lcal(H)
\end{equation}
instead of $\Lcal_\gamma(H)$.
Note that this description is  {\bf different} from the classical MAB  \cite[section 2.8]{sutton_reinforcement_2018} by the presence of the regularization term $\frac{\gamma}{2} \|H\|^2$. 
To solve 
\eqref{eq:problem_mab_as_maximization} the policy gradient 
approach prescribes the use of a gradient ascent stochastic algorithm which can be written~:
\begin{equation}
H_{t+1}(a) =H_t(a) + \rho_t  \left[ 
(R_t- \bar{R}_t) (\one_{a=A_t}- \Pi_{H_t}(a))- \gamma H_t(a)
\right]
, \ a=1,..., k,
\label{eq:definition_Ht_pg}
\end{equation}
where $R_t$ is the reward at time $t$, $\bar{R}_t$ is the mean reward up to time $t$ and $\rho_t$ a time step or 'learning rate' (see next section for the precise choice of time scheduling).
Although the formula \eqref{eq:definition_Ht_pg} seems somehow far from a stochastic gradient applied to $\Lcal$ we recall that this is indeed the case in the lemma~\ref{lemma:unbiased_grad} below.

\section{Theoretical convergence results} \label{sec:cv_proof}
We first recall why the term multiplying $\rho_t$ in 
 the right hand side of equation~\eqref{eq:definition_Ht_pg}
is indeed an unbiased estimation of $\nabla_H \Lcal(H)$.

To do this we need to be careful with the probabilistic framework; consider the filtration  $\Fcal_t$ corresponding to all information available up to time $t$. 
To go to $t+1$ two things happen: first the arm $A_t$ is
sampled with the discrete distribution $\Pi_{H_t}$; then a reward is sampled from the
distribution $R(A_t)$ of the arm $A_t$. As we will need very detailed information on this sampling, we need to make clear what part of the sampling is independent of $\Fcal_t$ and what part is measurable. 
Of course, $A_i$, $i<t$ and 
$H_i$, $i\le t$ are $\Fcal_t$ mesurables; but, since $A_t$'s distribution depend on $H_t$ it cannot be independent of $\Fcal_t$ as random variable. 
Nevertheless, in MAB sampling~: 
\begin{equation}
\E[\one_{a=A_t}|\Fcal_t] = \Pi_{H_t}(a).    
\label{eq:filtration_ft_at}
\end{equation}
To explain such a relation, imagine that the operations  at time $t$ start with  sampling some uniform variable $U_t$ in $[0,1)$  
independent of $\Fcal_t$ and then, depending on the value of $U_t$ a comparison is made with components of $\Pi_{H_t}$ to decide what value $A_t$ will take; this can be written
$\{A_t=a\} = \{ U_t \in [\sum_{b=1}^{a-1} \Pi_{H_t}(b), \sum_{b=1}^a \Pi_{H_t}(b)) \}$ with convention that the first sum is $0$ when $a=1$.
This gives equation \eqref{eq:filtration_ft_at}.
Now, once $A_t$ is chosen, the choice of the reward follows the same path: there is a part that is independent of the specific value of $A_t$, for instance one can draw another $V_t$ uniform in $[0,1]$ and attribute the reward based on the quantile of the $A_t$ distribution.

We denote 
\begin{equation}
 q_*(a) = \E [ R(a) ], \forall a \le k
\label{eq:hyp_mean0}
\end{equation}
which, considering the definition of $R_t$, means that 
\begin{equation}
\E [ R_t \one_{a=A_t}| \Fcal_t ] = q_*(a)\Pi_{H_t}(a), \forall a \le k.
\label{eq:hyp_mean}
\end{equation}
The following
hypothesis will be considered true from now on~:
\begin{equation}
\text{ there exists a constant } C_m>0 \text{ such that~: }
\E [ R(a)^2 ] \le C_m, \forall a \le k.
\label{eq:hyp_second_moment}
\end{equation}
\noindent 
We also introduce some notations for the terms appearing in the right hand side of \eqref{eq:definition_Ht_pg}~:
\begin{equation}
u_t(a):=(R_t- \bar{R}_t) (\one_{a=A_t}- \Pi_{H_t}(a)),
\label{eq:def_non_biased_gradient0}
\end{equation}
\begin{equation}
g_t(a):=(R_t- \bar{R}_t) (\one_{a=A_t}- \Pi_{H_t}(a))- \gamma H_t(a).
\label{eq:def_non_biased_gradient}
\end{equation}

We first give a preliminary result which explains why the algorithm \eqref{eq:definition_Ht_pg}
fits within the general framework of Robbins and Monro \cite{robbins_stochastic_1951}.
\begin{lemma} Under hypotheses \eqref{eq:hyp_mean} and 
\eqref{eq:hyp_second_moment}~:
\begin{equation}
\E    \left[ \left. g_t
\right|\Fcal_t \right] =   \left. \nabla_H \Lcal(H)\right|_{H=H_t}.
\label{eq:non_biased_gradient}
\end{equation}
Moreover, for some constant 
$C_{q_*}$ only depending on $q_*$ and $C_m$~:
\begin{equation}
\E [ \|g_t \|^2 ] \le C_{q_*} + 2 \gamma^2 \|H_t\|^2.
\label{eq:bounded_grad}
\end{equation}
\label{lemma:unbiased_grad}
\end{lemma}
\begin{remark}
The relation  \eqref{eq:non_biased_gradient}
says in essence that  \eqref{eq:definition_Ht_pg}
is  a Robbins-Monro type stochastic gradient in the sense that the 
stochastic estimate $g_t$ of the gradient $\left.\nabla_H \Lcal(H)(a)\right|_{H=H_t}$ is unbiased.
On the contrary, \eqref{eq:bounded_grad} is a 
technical point that will be required latter.
\end{remark}
\begin{proof}
\noindent
{\bf  Equality \eqref{eq:non_biased_gradient}~:}
Of course, the gradient of the $L2$ regularization term $\frac{\gamma}{2} \|H\|^2$ is $\gamma H$ which explains its presence in the left hand side, i.e., in $g_t$. 
On the other hand, the baseline $\bar{R}_t$ satisfies~: 
\begin{equation}
 \E  [\bar{R}_t (\one_{a=A_t}- \Pi_{H_t}(a))|\Fcal_t ] = 
\bar{R}_t \cdot \Pbb[A_t = a] - \bar{R}_t  \Pi_{H_t}(a)= 0.   
\end{equation}
Only the gradient of the reward $R$ remains to be computed;  we proceed as in \cite[Section 2.8]{sutton_reinforcement_2018} by 
recalling that from \eqref{eq:hyp_mean} it follows that 
$\E[R_t | \Fcal_t] =\E[ \sum_a R(a) \one_{a=A_t} | \Fcal_t] = \sum_a q_*(a) \Pi_{H_t}(a)$ which
implies 
\begin{equation}
 \Lcal(H)=\langle q_*, \Pi_H \rangle -\frac{\gamma}{2} \|H\|^2.
\label{eq:L_as_scalar_prod}
\end{equation}
To conclude, it is enough to invoke the formula of the derivatives of the softmax function $H \mapsto \Pi_H$~:
\begin{equation}{\frac{\partial \Pi _H(a)}{\partial H(b)}}=\Pi_H(a)\left( \one_{a=b}-\Pi_H(b)\right ).
\label{eq:softmax_derivative}
\end{equation}

\noindent
{\bf  Estimation \eqref{eq:bounded_grad}~:} since 
$g_t = u_t - \gamma H_t$, we only have to prove a bound for 
$\E[\|u_t\|^2 | \Fcal_t]$. First note that 
$|\one_{a=A_t}-\Pi_{H_t}(a)| \le 1$ so we are left with finding a bound for $\E[\|R_t - \bar{R}_t\|^2]$; but 
from \eqref{eq:hyp_second_moment}~: 
\begin{equation}
\E[\|R_t - \bar{R}_t\|^2] \le 2 \E[\|R_t\|^2] +2\E[\|\bar{R}_t\|^2] \le 2 C_m +2\E[\|\bar{R}_t\|^2].    
\end{equation}
On the other hand $\bar{R}_t = \frac{R_0+\cdots + R_{t-1}}{t}$ with all terms having bounded second order moment (by \eqref{eq:hyp_second_moment}) which shows that $\E[\|\bar{R}_t\|^2] \le C_m$ hence the conclusion.
\qed\end{proof}

\subsection{Fixed time step}
We prove now the first result involving the $L2$ regularized MAB including the case when the time step is constant (but small enough to ensure convergence) and $\gamma$ large enough.
\begin{proposition}
Denote 
\begin{equation}
\mu := \gamma -  (\max_a q_*(a)- \min_a q_*(a)).    
\label{eq:def_mu}
\end{equation}
 Under the hypotheses \eqref{eq:hyp_mean} and 
\eqref{eq:hyp_second_moment} assume 
\begin{equation}
    \mu >0. 
\end{equation}
Then~: 
 \begin{enumerate}
\item the function $\Lcal$ defined in \eqref{eq:functional_to_min} has a unique maximum $H_*$;
\label{item:uniqueness}
     \item \label{item:inequalitydn}
For any $t \ge 0$ denote
 \begin{equation}
     d_t =  \E \left[ \|H_{t}-H_*\|^2 \right].
 \end{equation}
 Then there exist constants $c_2, c_3,c_4 >0$ depending only on $q_*$ such that for $c_1 = c_4 \gamma^2$, $c_0=c_2 + c_3\gamma^2$~:
 \begin{equation}
     d_{t+1} \le      (1-\rho_t \mu + \rho_t^2 c_1) d_t + \rho_t^2 c_0.
\label{eq:dnplusone} \end{equation}
\item \label{item:constant}
For any $\epsilon > 0$ there exists a $\rho_\epsilon > 0$ such that if $\rho_t = \rho < \rho_\epsilon$ then 
\begin{equation}
    \limsup_{t\to \infty}  \E \left[ \|H_{t+1}-H_*\|^2 \right] \le \epsilon.
\label{eq:estimationvoisinage}
\end{equation}
\item \label{item:convergence}
Take $\rho_t$ a sequence such that:
\begin{equation}
\rho_t \to 0 \text{ and } \sum_{t\ge 1} \rho_t = \infty.
\label{eq:hyprhon}
\end{equation}
 Then $d_t \to 0$, or equivalently 
\begin{equation}
 \lim_{t \to \infty }H_t \overset{\mathrm{L^2}}{=} H_*.
\end{equation}
\end{enumerate}
\label{prop:rho_constant}
\end{proposition}
\begin{proof}
{\bf Item \ref{item:uniqueness}}:
We first establish some estimates concerning the Hessian $\nabla_H ^2 \Lcal$; Take $c$ to be a constant. We can write~:
\begin{equation}\begin{array}{ll}
\nabla_H ^2 \Lcal(H) = \nabla_H ^2 (\Lcal(H)-c) & = 
\nabla_H ^2 \left( \langle q_*-c,\Pi_H\rangle - \frac{\gamma}{2} \|H \|^2\right) \\
~& = \nabla_H ^2 \left( \langle q_*-c,\Pi_H\rangle \right) - \gamma I_k.
\end{array}\end{equation}
On the other hand, if we iterate the equation 
\eqref{eq:softmax_derivative}
once more we obtain for $A,a,b \le k$~:
\begin{equation}\begin{array}{ll}
\displaystyle \frac{\partial ^2\Pi_H(A)}{\partial H(b)\partial H(a)}
& =\Pi_H(A)\left( \one_{a=A}-\Pi_H(a)\right) \left( \one_{b=A}-\Pi_H(b)\right) \\
~& \ \  -\Pi_H(A)\Pi_H(a)\left( \one_{b=a}-\Pi_H(b)\right) \le 2\Pi_H(A).
\label{eq:softmax_second_derivative_}
\end{array}\end{equation}
From this we obtain for any $\bar{H}$ and variations $\delta H$~:
\begin{eqnarray}
& \ & 
\nabla^2_H \langle q_*-c,\Pi_H\rangle|_{H=\overline{H}} (\delta H,\delta H) \nonumber \\ & \ &
=\sum_{A=1}^k (q_*(A)-c)\sum_{a,b=1}^k{\frac{\partial ^2\Pi_H(A)}{\partial H(b)\partial H(a)}}\Big|_{H=\overline{H}}\delta H(a)\delta H(b) 
\nonumber \\ & \ & 
= \sum_{A=1}^k (q_*(A)-c) \Pi_{\overline{H}}(A) [
\langle \delta H(A)-\delta H,\Pi_{\overline{H}} \rangle^2
- \langle \delta H^2,\Pi_{\overline{H}} \rangle + \langle \delta H,\Pi_{\overline{H}} \rangle^2 ] \nonumber \\ & \ &  
\le \max_A |q_*(A)-c|\cdot
|\langle \delta H(A)-\delta H,\Pi_{\overline{H}} \rangle^2
- \langle \delta H^2,\Pi_{\overline{H}} \rangle + \langle \delta H,\Pi_{\overline{H}} \rangle^2|.
\end{eqnarray}
Take now $c= \frac{\max_a q_*(a) + \min_a q_*(a)}{2}$; then
\begin{equation}
\max_a|q_*(a)-c| = \frac{\max_a q_*(a) - \min_a q_*(a)}{2}=: \frac{c_*}{2} ,  
\label{eq:notation_cstar}
\end{equation}
where the 
second part is a notation; since by Cauchy 
$\langle \delta H^2,\Pi_{\bar{H}} \rangle - \langle \delta H,\Pi_{\bar{H}} \rangle^2\ge 0$
the term
$\langle \delta H(A)-\delta H,\Pi_{\bar{H}} \rangle^2
- \langle \delta H^2,\Pi_{\bar{H}} \rangle + \langle \delta H,\Pi_{\bar{H}} \rangle^2$
is the difference of two positive numbers so its absolute value is smaller than the largest of them. We will prove that each is smaller than $2 \|\delta H\|^2$.
Obviously 
$\langle \delta H^2,\Pi_{\bar{H}} \rangle - \langle \delta H,\Pi_{\bar{H}} \rangle^2\le 
\langle \delta H^2,\Pi_{\bar{H}} \rangle\le \max_a \delta H(a)^2 \le \|\delta H\|^2$.
For the first term we look for an optimum of $\langle \delta H(A)-\delta H,\Pi_{\bar{H}} \rangle^2$ under 
the constraint $\|\delta H \|^2=1$ and, after some straightforward computations we obtain $2$ (see Lemma~\ref{lemma:max_hess_contraint} for a proof).
Thus finally~:
\begin{eqnarray}
& \ & 
\nabla^2_H \langle q_*-c,\Pi_H\rangle|_{H=\overline{H}} (\delta H,\delta H) \le c_* \|\delta H\|^2.
\end{eqnarray}
It follows from the previous considerations that \begin{equation}
\nabla^2_H \Lcal(H)|_{H=\overline{H}} (\delta H,\delta H) \le 
(c_*-\gamma) \|\delta H\|^2.
\label{eq:estimate_hessianL}
\end{equation}

Take $H \in \R^k$. Using Taylor's formula for $s \mapsto sH$ we obtain some 
$\bar{H}$ on the segment $[0,H]$ such that~:
\begin{eqnarray}
& \ & 
\Lcal(H)=\Lcal(0) + \langle \nabla_H \Lcal(0), H \rangle + 
\frac{1}{2}\nabla^2_H \Lcal(H)|_{H=\overline{H}} (H,H) 
\nonumber \\ & \ & 
\qquad \le 
\Lcal(0) + \langle \nabla_H \Lcal(0), H \rangle + (c_*/2-\gamma/2) \| H\|^2.
\end{eqnarray}
When $\gamma > c_*$ we obtain that $-\Lcal$ is coercive at infinity thus by continuity we obtain the existence of an optimum. The uniqueness follows from the 
strict concavity of $\Lcal$ (see inequality \eqref{eq:estimate_hessianL}).

\noindent {\bf Item \ref{item:inequalitydn}}:
We have
\begin{align}
&    \E \left[ \|H_{t+1}-H_*\|^2 \right]  = 
    \E \left[ \|H_{t}-H_* + \rho_t g_t\|^2 \right]  
\nonumber    \\& \quad = 
    \E \left[ \|H_{t}-H_*\|^2 \right]  +
\rho_t^2    \E \left[ \|g_t\|^2 \right]  
+ 2 \rho_t 
    \E \left[ \langle H_{t}-H_*, g_t\rangle \right].
\end{align}
From \eqref{eq:non_biased_gradient}
$$   \E \left[ \langle H_{t}-H_*, g_t\rangle \right]
 = \E \left[ \langle H_{t}-H_*, \nabla_H \Lcal(H_t)\rangle \right].
$$
Recall that since $H_*$ is an optimum $\Lcal(H_*)\ge\Lcal(H_t)$; 
using a Taylor expansion for $s\mapsto s H_* + (1-s) H_t$ around $H_t$ and using the same estimations as above for the Hessian we obtain
\begin{align}
\E \left[ \langle H_{t}-H_*, \nabla_H \Lcal(H_t)\rangle \right] & \le 
\E \left[ \Lcal(H_t) - \Lcal(H_*) - \frac{\mu}{2}  \|H_{t}-H_*\|^2 \right] 
\nonumber\\~ & 
\le - \frac{\mu}{2}  \E  [\|H_{t}-H_*\|^2].
\label{eq:estim_grad_proof2}
\end{align}
Combining 
all these estimations and using \eqref{eq:bounded_grad} 
 to bound the term $\E [\|g_t\|^2]$
we obtain the inequality \eqref{eq:dnplusone}.
For the rest of the proof we follow the proof of Thm. 1 in \cite{gabriel_turinici_convergence_2023}. 

\noindent {\bf Item \ref{item:constant}}:
When $\rho_t=\rho$ estimation \eqref{eq:dnplusone} is written 
$$d_{t+1} -  \frac{\rho c_0}{\mu- \rho c_1} \le (1-\rho \mu+ \rho^2 c_1) \left(d_t - \frac{\rho c_0}{\mu- \rho c_1}\right).$$
If $\rho < \min(1/\mu, \mu/2 c_1)$, taking the positive part allows to write~:
$$\left(d_{t+1} - \frac{\rho c_0}{\mu- \rho c_1} \right)_+
\le \left(1- \frac{\rho \mu}{2}\right) \left(d_t - \frac{\rho c_0}{\mu- \rho c_1}\right)_+,$$
and therefore $\forall \ell\ge 1$:
$$\left(d_{n+\ell} - \frac{\rho c_0}{\mu- \rho c_1} \right)_+
\le \left(1-\frac{\rho \mu}{2}\right)^\ell \left(d_t - \frac{\rho c_0}{\mu- \rho c_1}\right)_+.$$
For $\ell\to \infty$ we obtain 
$\limsup_\ell \left(d_{\ell} - \frac{\rho c_0}{\mu- \rho c_1} \right)_+=0$ which gives the conclusion \eqref{eq:estimationvoisinage} for $\rho \le \rho_\epsilon:=\min\{1/\mu, \mu/2c_1, \epsilon \mu / (c_0+\epsilon c_1)\}$.

\noindent {\bf Item \ref{item:convergence}}:
Consider now $\rho_t$  non-constant and fix $\epsilon >0$; 
we invoke inequality \eqref{eq:dnplusone} and obtain~:
$$d_{t+1} - \epsilon \le \left(1-\frac{\rho_t \mu}{2}\right) (d_t - \epsilon) + \rho_t (c_0\rho_t - \mu \epsilon/2 + (\rho_t c_1 - \mu/2)d_t).$$ 
When $t$ is big enough, the last term in the right hand side is negative and 
therefore $$d_{t+1} - \epsilon \le \left(1-\frac{\rho_t \mu}{2}\right) (d_t - \epsilon),$$ hence 
$$\left(d_{t+1} - \epsilon\right)_+
\le \left(1-\frac{\rho_t \mu}{2}\right) \left(d_t - \epsilon\right)_+.$$
Taking the product of all relations of this type allows to write~:
\begin{equation}
\left(d_{t+\ell} - \epsilon \right)_+
\le \prod_{s=t}^{t+\ell-1} \left(1-\frac{\rho_s \mu}{2}\right) \left(d_t - \epsilon \right)_+.    
\label{eq:upperbound2}
\end{equation}
Using the Lemma 2  from \cite{gabriel_turinici_convergence_2023} recalled 
as Lemma \ref{lemma:product} below 
we obtain
$\lim_{\ell \to \infty} \left(d_{\ell} - \epsilon \right)_+ = 0$ and since this is true for any $\epsilon$ the conclusion follows.
\qed\end{proof}

\begin{lemma} Let $\Pi\in \R^k$, $\Pi(a) \ge 0$, $\forall a \le k$, $\sum_a\Pi(a)=1$. Then for any 
$x \in \R^k$ and any $\ell \le k$
\begin{equation}
    (\langle x,\Pi\rangle - x_\ell )^2 \le 2 \|x\|^2.
\end{equation}
\label{lemma:max_hess_contraint}
\end{lemma}
\begin{proof}
The term $\langle x,\Pi\rangle$ is a mean value of $x$  under the law $\Pi$ thus it is somewhere
between the smallest (denoted $x_m$) and the largest (denoted $x_M$) values of $x_i$, $i\le k$.
The left hand side is thus smaller than $(x_M-x_m)^2$. On the other hand,
$2 \|x\|^2 \ge 2(x_m^2 + x_M^2) \ge (x_M-x_m)^2$. 
\qed\end{proof}

\subsection{Convergence rates for linear decay  $\rho_t= \frac{\beta_1}{1+\beta_2 t }$ and large $\gamma$}
We investigate now the situation when $\rho_t$ is not
constant but  decays linearly.
 \begin{proposition} 
Let $\beta_1, \beta_2 > 0$ two positive constants and take
\begin{equation}
\rho_t= \frac{\beta_1}{1+\beta_2 t }.    
\label{eq:rhot_beta_def}
\end{equation}
 Under the hypotheses \eqref{eq:hyp_mean} and 
\eqref{eq:hyp_second_moment} for $\gamma$ large enough~:
\begin{enumerate}
\item  the problem \eqref{eq:problem_mab_as_maximization}  has a unique solution $H_*$;
\item the $L2$ regularized policy gradient MAB algorithm
\eqref{eq:definition_Ht_pg}
converges with the rate~:
\begin{equation}\mathbb{E}[ \|H_t-H_*\|^2]= O\left(  \frac{1}{t}\right)
\textrm{ as } t\to \infty.
\label{eq:order_cv_mab_sgd}
\end{equation}
\end{enumerate} \label{prop:cv_rate}
\end{proposition}
\begin{proof}
We saw already in proposition~\ref{prop:rho_constant}
that the optimum exists and is unique for $\mu>0$.
We also saw that 
\eqref{eq:dnplusone} is satisfied.
Denote $\xi_t = t d_t$.
By multiplication with $t+1$ inequality \eqref{eq:dnplusone} can be written in terms of $\xi$ as
\begin{equation}
\xi_{t+1} \le (1-\rho_t \mu + c_1 \rho_t^2) \xi_t (1+\frac{1}{t}) + c_0 \rho_t^2 (t+1).
\label{eq:xi_ineq}
\end{equation}
It is enough to prove that $\xi_t$ is bounded to conclude. Suppose on the contrary that $\xi_t$ is not bounded. In this case, for any $C$ large enough there exists some rank $t_C$ large enough where $\xi_{t+1}$
is for the first time larger than $C$. In particular this means that $\xi_t \le C \le \xi_{t+1}$. Therefore
\begin{equation}
 C \le \xi_{t+1} \le  (1-\rho_t \mu + c_1 \rho_t^2) C (1+\frac{1}{t}) + c_0 \rho_t^2 (t+1).
\end{equation}
so finally 
\begin{equation}
 C \le  (1-\rho_t \mu + c_1 \rho_t^2) C (1+\frac{1}{t}) + c_0 \rho_t^2 (t+1),
\end{equation}
 or, after simplification by $C$ in both terms and multiplication by $t$:
\begin{equation}
0 \le  C \left[ t(-\rho_t \mu + c_1 \rho_t^2) + 1 -\rho_t \mu + c_1 \rho_t^2\right]  + c_0 \rho_t^2 (t+1)t.
\label{eq:C_ineq}
\end{equation}
Recall that $\rho_t = \frac{\beta_1}{1+\beta_2 t}$. For $t\to \infty$ the term
$c_0 \rho_t^2 (t+1)t$ tends to the constant $c_0 \frac{\beta_1^2}{\beta_2^2}$.
The terms multiplying $C$ tends to 
\begin{equation}
    \lim_{t\to \infty}  \left[ t(-\rho_t \mu + c_1 \rho_t^2) + 1 -\rho_t \mu + c_1 \rho_t^2\right]  = 
1-\mu \frac{\beta_1}{\beta_2}.
\end{equation}
But for $\mu$ large enough (i.e., $\gamma$ large enough) this is negative so the right hand side in 
\eqref{eq:C_ineq} cannot remain positive when $t$ and $C$ are large enough because is a sum of a bounded term and a product between $C$ and a quantity that converges to a strictly negative constant. This provides the required contradiction and ends the proof.
\qed\end{proof}

\subsection{Behavior when $\gamma\to 0$}
For completeness, we investigate in this section the other side of the question, namely the regularization part. 

The real goal is to solve problem 
\eqref{eq:problem_mab_as_maximization} for $\gamma=0$. 
When $\gamma >0$ the solution of the problem 
\eqref{eq:problem_mab_as_maximization}
will not coincide with the solution for $\gamma=0$. The question is whether this perturbation will be small when $\gamma$ is small. This is an intuitive result but not completely trivial because when $\gamma=0$ the maximum in 
$\eqref{eq:problem_mab_as_maximization}$ is not attained in general as  most of its components will tend to $-\infty$. Indeed, suppose all $q_*(a)$ are different and denote by 
$q_*(a_{max})$ the largest one. When $\gamma=0$ the functional is simply
$\Lcal_0(H)= \sum_a q_*(a) \Pi_H(a) < q_*(a_{max})$. The inequality is always strict but 
$q_*(a_{max})-\Lcal_0(H)$
vanishes when $\Pi_H$ is a Dirac mass in 
$a_{max}$; for that to happen $H$ would have to have all entries equal to $-\infty$ except
$H_{a_{max}}$ that can be any finite value.

The result below informs that, as expected, we can be as close as we want to the optimum value of the non-regularized MAB problem by taking $\gamma$ small enough.
\begin{lemma}
Let 
\begin{equation}
    V(\gamma) := \max_{H \in \R^k} \Lcal_\gamma(H).
\end{equation}
Then
$\lim_{\gamma \to 0} V(\gamma) = V(0)$.
\label{lemma:cv_gamma_to_zero}
\end{lemma}
\begin{proof}
For any $\gamma >0$
denote $H_\gamma^*$ one optimum in \eqref{eq:problem_mab_as_maximization}.
Note first that, by standard coercivity and continuity on compacts arguments of $-\Lcal_\gamma$,
this optimum value exists for any $\gamma>0$ 
(it is not necessarily unique though);  for $\gamma=0$ it does not exist in general as the maximum value is attained only as a limit of values along a sequence $H_n$. Take $H_n$ to be a sequence such that  
\begin{equation}
  \lim_{n\to \infty}  \Lcal_0(H_n) = \sup_{H \in \R^k} \Lcal_0(H) = V(0).
\end{equation}
By the very definition of $H_\gamma^*$~:
\begin{equation}
V(\gamma)= \Lcal_\gamma(H_\gamma^*) \ge \Lcal_\gamma(H), \ \forall H \in \R^k.
\end{equation}
In particular 
\begin{equation} V(\gamma)= 
\Lcal_\gamma(H_\gamma^*) \ge \Lcal_\gamma(H_n) = \Lcal_0(H_n) - \frac{\gamma}{2} \|H_n\|^2,\  \forall n \ge 1 .
\end{equation}
Keep now $n$ fixed and let $\gamma\to 0$ we obtain $\liminf_{\gamma\to 0}{V(\gamma)} \ge \Lcal_0(H_n)$. Take now $n\to \infty$ to obtain
 $V(0) \le \liminf_{\gamma\to 0}{V(\gamma)}$. 
 But since on the other hand for all $\gamma \ge 0$~: $V(\gamma) \le V(0)$  we obtain the conclusion.
\qed\end{proof}

\section{Numerical simulations} \label{sec:numerical}

\begin{figure*}
\begin{center}
\includegraphics[width=0.49\linewidth]{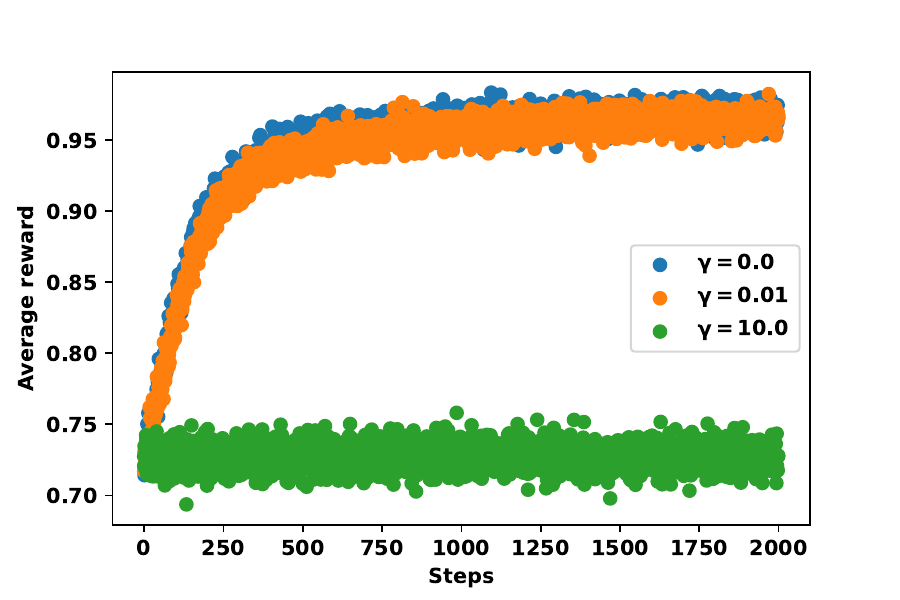}
\includegraphics[width=0.49\linewidth]{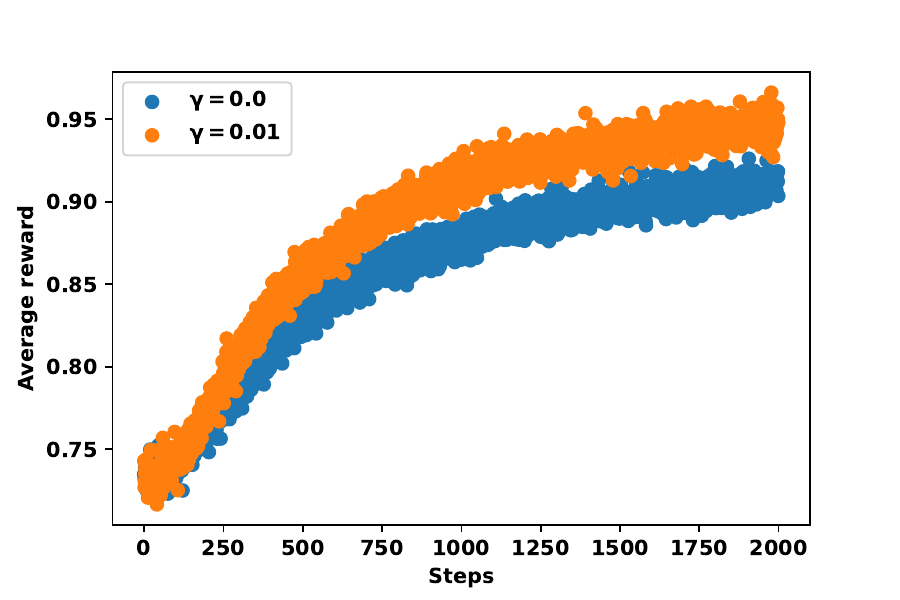}
\caption{The average reward for $\rho_t=0.05$ (constant), $\gamma$ is $0$, $0.01$ or $10$ (see the legend).
{\bf Left~:~}
 start from a uniform distribution 
$\Pi_{H_0}$ with
$H_0=(0,...,0)$. {\bf Right~:~}  start from a biased distribution 
$\Pi_{H_0}$ with
 $H_0=(5,...,0)$.}  \label{fig:biased_and_not_biased}
\end{center}
\end{figure*}

The Python implementation is available on Github
~\footnote{\scriptsize\url{https://github.com/gabriel-turinici/regularized_policy_gradient} version August 31st 2024.}.
We perform 
$M=1000$ tests of $2000$ steps each; for each of the $M$ tests we sample, as in \cite[fig 2.5 page 38]{sutton_reinforcement_2018}
$k=10$ arms with 
$q_*(a)$, $a=1, ..., k$ independent and normally distributed with mean $4$ and unit variance.

Once $q_*(\cdot)$ have been sampled they do not change for the $2000$ steps of the respective test. To ensure fair comparison we use same values of $q_*(\cdot)$ for all the bandits that are compared, for instance in figure~\ref{fig:biased_and_not_biased} run number 123 for $\gamma=0$ and run number 123 for $\gamma=0.01$ and 
run number 123 for $\gamma=10$ share the same  $q_*(\cdot)$, which is different from the  $q_*(\cdot)$ of runs 122.
For each of the arms $a=1, ..., k$ the  law of $R(A)$ conditional to $A=a$  is a normal 
variable with mean $q_*(a)$ and unit variance. Note that in this case 
\begin{equation}
c_*^{avg} := \E [\max_{a\le k} q_*(a)- \min_{a\le k} q_*(a)] \simeq 3.08.
\end{equation}
The proposition~\ref{prop:rho_constant} prescribes that $\gamma$ should be larger than
$c_*^{avg}$. 

The uniform distribution corresponding to $H_0=(0,...,0)$ would give, in average, a reward equal to $4$. Nevertheless in the following, for each of the $M$ tests  we will not plot the absolute value of the reward but the value relative to
the maximum possible reward $\max_{a\le k} q_*(a)$ (because this maximum varies with each run). With this convention the best possible reward is $1$.
The average over the $M=1000$ runs are presented in 
figure~\ref{fig:biased_and_not_biased}
and discussed also in section~\ref{sec:conclusion}. 
We see that when starting for the uniform distribution 
the regularization $\gamma=0.01$ does not prevent the algorithm to have a performance comparable with the non-regularized version (i.e., $\gamma=0$).
On the contrary, the value $\gamma=10$ is too large and biases the algorithm towards a non-optimal solution (similar results are obtained for $\gamma= 3.08$).
When starting from a biased distribution, the regularization $\gamma=0.01$ does a better job and obtains a visible improvement over the performance of the non-regularized version (i.e., $\gamma=0$).

\begin{figure*}
\begin{center}
\includegraphics[width=0.75\linewidth]{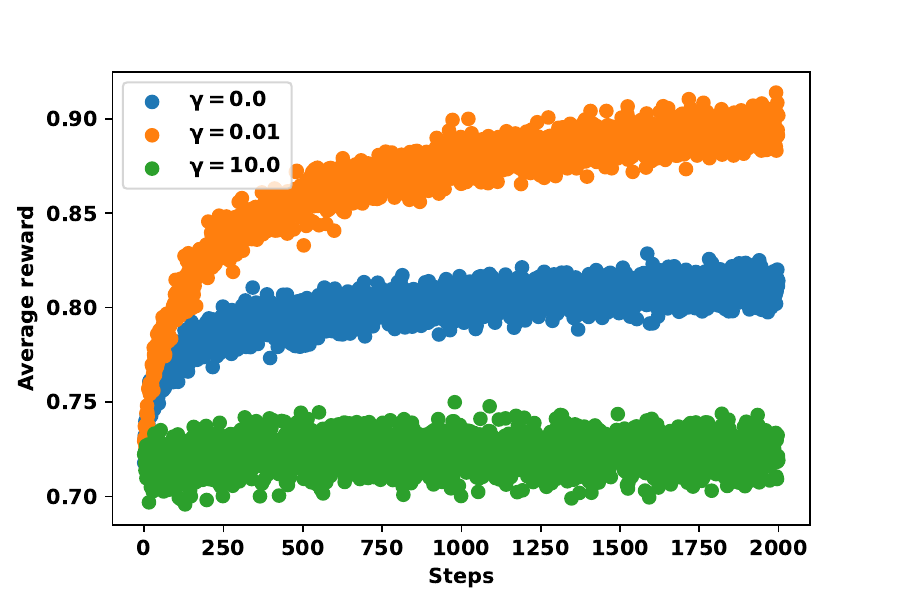}
\caption{
The average reward when starting from the non-uniform distribution 
$\Pi_{H_0}$ with
 $H_0=(5,...,0)$ and 
$\rho_t=\frac{1}{1+0.05*t}$ 
in the general setting of proposition \ref{prop:cv_rate} equation \eqref{eq:rhot_beta_def};
we test  $\gamma=0$, $\gamma=0.01$ or $\gamma=10$ (see the legend). 
As before, $\gamma=10$ is too large to obtain good results.} 
\label{fig:rhot_decay}
\end{center}
\end{figure*}

Additional tests are presented in figures \ref{fig:rhot_decay} and \ref{fig:gamma_not_constant} where we investigate a linear decay schedule for the learning rate $\rho_t$ and also for the regularization coefficient $\gamma_t$.
In particular in figure~\ref{fig:gamma_not_constant} it is seen that 
the non-null initial regularization helps leaving the non-optimal initial guess $H_0$; then the decay of $\gamma_t$ will provide results comparable the non-regularized version in particular convergence to an optimal point.

\section{Summary and discussion} \label{sec:conclusion}

We considered in this work a $L2$ regularized policy gradient algorithm applied to a Multi Armed Bandit (MAB) and investigated it both theoretically (convergence, rate of convergence) and numerically. Let us first recall that is was already remarked in the literature \cite{mei_global_2020} that the MAB may behave erratically when the initialization is close to a sub-optimal critical point (there are many of them, for instance all Dirac masses are critical points). In this case the standard gradient policy MAB will spend a long time in this  region before converging to the global maximum. One  way to cure this drawback is to introduce regularization, in our case this is $L2$ regularization, parameterized by a multiplicative coefficient  $\gamma$.

Under technical conditions on the value of  $\gamma$  we gave two convergence results~:  proposition~\ref{prop:rho_constant} that works for both constant and variable time steps $\rho_t$ and proposition~\ref{prop:cv_rate} that proves that the convergence happens at rate 
$O(1/t)$ if $\rho_t$ decays linearly. However the existence of the regularization part (when $\gamma>0$) may shift the optimal solution; we proved then in lemma~\ref{lemma:cv_gamma_to_zero} that when $\gamma\to 0$ the optimality is restored.

\begin{figure*}
\begin{center}
\includegraphics[width=0.75\linewidth]{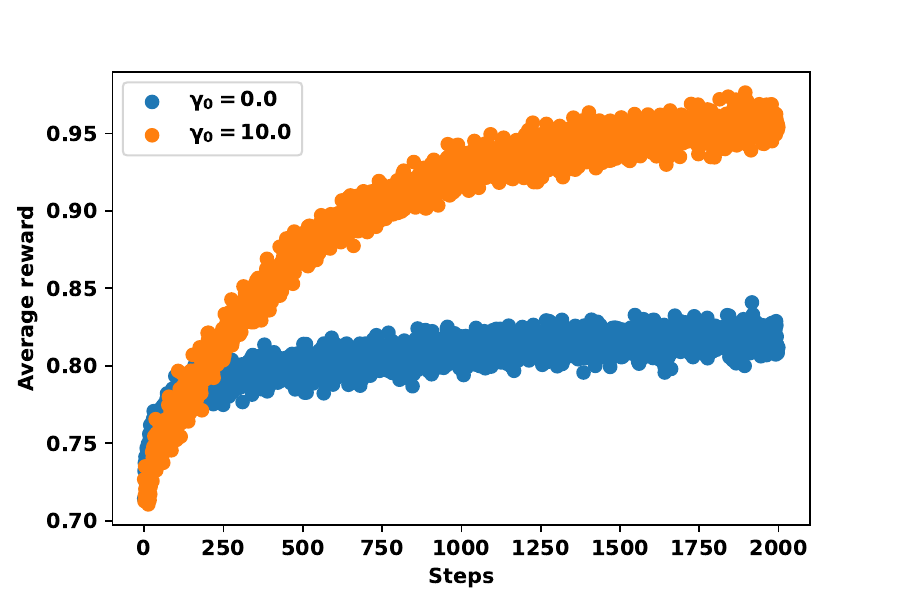}
\caption{
The  average reward when starting from the biased distribution 
$\Pi_{H_0}$ with
 $H_0=(5,...,0)$ and $\gamma_t= \frac{\gamma_0}{1+0.2*t}$ (see the legend), $\gamma_0=0$ (no regularization) or $\gamma_0=10$.
We take $\rho_t=\frac{1}{1+0.05*t}$ (see eq. \eqref{eq:rhot_beta_def}). 
}  \label{fig:gamma_not_constant}
\end{center}
\end{figure*}

The technical conditions in the theoretical results impose large values of $\gamma$ but in practice small values of $\gamma$ are requested for good quality solution. To see the usefulness of the regime when $\gamma$ is small, we tested the procedure numerically. The results indicate that
irrespective of whether $\gamma$ is large or small the convergence occurs when the initial guess $H_0$ is uniform, but the quality of the optimum is not good when $\gamma$ is too large, see figure~\ref{fig:biased_and_not_biased};
    same holds true when non-constant (linear decay) $\rho_t$ is used, see figure~\ref{fig:rhot_decay};
    when the initial guess $H_0$ is {\bf not uniform} the convergence is significantly better with $\gamma >0$ and for $\gamma$ not too large its quality is also very good, see figure~\ref{fig:biased_and_not_biased}.

To combine the best of the two possible worlds, we also tested a variable $\gamma_t$ of the form $\gamma_t = \frac{\gamma_0}{1+\eta t}$ (with $\eta$ a positive constant) 
starting from non-uniform initial guess $H_0$; the results in
 figure~\ref{fig:gamma_not_constant}, are very good and show that this choice is better than the classical non regularized policy gradient procedure. The precise optimal decay schedule for $\gamma_t$ is not known and will be left for future works.
\appendix
\section{Appendix}
For completeness we recall  below the Lemma 2 from \cite{gabriel_turinici_convergence_2023} and its proof.
\begin{lemma}
Let $\xi >0$ and $\rho_t$ a sequence of positive real numbers such that $\rho_t \to 0$ and $\sum_{t\ge 1} \rho_t = \infty$. Then for any $t \ge 0$:
\begin{equation}
\lim_{\ell\to \infty}    \prod_{j= t}^{t+\ell} (1-\rho_j \xi) = 0.
\end{equation}
\label{lemma:product}
\end{lemma}
\begin{proof}
Since $\rho_t\to 0$,  $\rho_j \xi<1$ for $j$ large enough; without loss of generality  we can suppose this is true starting from $t$. 
Since for any $x \in ]0,1[$ we have $\log(1-x) \le -x$~:
\begin{equation}
0 \le   \prod_{j= t}^{t+\ell} (1-\rho_j \xi) = e^{ \sum_{j= t}^{t+\ell} \log(1-\rho_j \xi)} \le e^{ \sum_{j=t}^{t+\ell} (-\rho_j \xi)} \overset{\ell\to \infty}{\longrightarrow} e^{- \infty} =0.
\end{equation}
\qed\end{proof}

\section{Further comments on the assumption $\mu>0$}
The convergence of the scheme proved in proposition~\ref{prop:rho_constant} requires $\mu :=\gamma -c_*>0$ ($c_*$ is defined in \eqref{eq:notation_cstar}). If for some reason a $\gamma$ is given and cannot be changed and $\gamma -c_* \le 0$ 
we can still get convergence if we consider the modified softmax parametrized MAB policy gradient algorithm that finds $H\in \mathbb{R}^k$ solution to 
\eqref{eq:problem_mab_as_maximization}
where the functional $\Lcal_\gamma$ is replaced by~:
$\mathbb{E}_{A \sim \Pi^\alpha_H} \left[ R(A) - \frac{\gamma}{2}\| H\|^2\right]$
where
$\Pi^\alpha_H(A):=\frac{e^{\alpha H(A)}}{\sum_{a=1}^ke^{\alpha H(a)}}$ 
i.e., $\Pi^\alpha_H = \Pi_{\alpha H}$.    
Here $\alpha >0$ is an arbitrary but fixed constant
such that
$\gamma -\alpha ^2c_*>0$. Note that the only change is the replacement of 
$\Pi_H$ by $\Pi^\alpha_H$. With this provision the stochastic gradient ascent  
algorithm \eqref{eq:definition_Ht_pg} is replaced by~:
\begin{equation}
H_{t+1}(a)=H_t(a)+\rho _t\Big[ \alpha (R_t-\bar{R}_t)(\one_{a=A_t}-\Pi^\alpha_{H_t}(a))-\gamma H_t(a)\Big] \, , \ a=1, ..., k.    
\label{eq:definition_Ht_pg_alpha}
\end{equation}
The proof of the Proposition~\ref{prop:rho_constant} remains the same as soon as we  replace $\mu$ in \eqref{eq:def_mu} with
$\gamma -\alpha ^2c_*>0$  and work with 
\begin{equation}
u_t:=\alpha (R_t-\bar{R}_t)(\one_{a=A_t}-\Pi^\alpha_{H_t}(a))
\end{equation}
\begin{equation}
g_t:=\alpha (R_t-\bar{R}_t)(\one_{a=A_t}-\Pi^\alpha_{H_t}(a))-\gamma H_t(a).\end{equation}
The proofs of Lemma \ref{lemma:unbiased_grad}
 and Proposition \ref{prop:rho_constant}  will use 
\begin{equation}
\frac{\partial \Pi^\alpha_H(a)}{\partial H(b)}=\alpha \Pi^\alpha_{H}(a)(\one_{a=b}-\Pi^\alpha_H(b))\end{equation}
and
\begin{equation*}
\frac{\partial ^2\Pi^\alpha_H(A)}{\partial H(b)\partial H(a)}=\alpha ^2\Pi^\alpha_H(A)(\one_{a=A}-\Pi^\alpha_H(a))(\one_{b=A}-\Pi^\alpha_H(b))\end{equation*}
\begin{equation}
-\alpha ^2\Pi^\alpha_H(A)\Pi^\alpha_H(a)(\one_{b=a}-\Pi^\alpha_H(b))\leq 2\alpha ^2\Pi^\alpha_H(A).\end{equation}
Note however that 
for  given $\alpha, \gamma$
this modified procedure 
will converge to the same optimal $H_*$ as the original procedure 
(i.e. $\alpha=1$) 
if we take $\gamma/\alpha^2$ instead of the original $\gamma$ (direct computations show that they share the same critical point equations).


\end{document}